\documentclass{article}

\usepackage{microtype}
\usepackage{graphicx}
\usepackage{subcaption}
\usepackage{booktabs} 

\usepackage{hyperref}

\usepackage[accepted]{icml2025}

\usepackage{amsmath}
\usepackage{amssymb}
\usepackage{mathtools}
\usepackage{amsthm}

\usepackage[capitalize,noabbrev]{cleveref}

\theoremstyle{plain}

\theoremstyle{definition}

\theoremstyle{remark}

\usepackage[textsize=tiny]{todonotes}

\usepackage{amssymb}
\usepackage{pifont}
\newcommand{\cmark}{\ding{51}}%
\newcommand{\xmark}{\ding{55}}%
\usepackage{multirow, subcaption}
\usepackage[table]{xcolor}

\icmltitlerunning{Instruction Tuning of Large Language Models for Tabular Data Generation---in One Day}

\begin{document}

\twocolumn[
\icmltitle{Instruction Tuning of Large Language Models \\ for Tabular Data Generation---in One Day}




\begin{icmlauthorlist}
\icmlauthor{Milad Abdollahzadeh}{SIT,betterdata}
\icmlauthor{Abdul Raheem}{betterdata}
\icmlauthor{Zilong Zhao}{NUS,betterdata}
\icmlauthor{Uzair Javaid}{betterdata}
\icmlauthor{Kevin Yee}{betterdata}
\icmlauthor{Nalam Venkata Abhishek}{SIT}
\icmlauthor{Tram Truong-Huu}{SIT}
\icmlauthor{Biplab Sikdar}{NUS}
\end{icmlauthorlist}

\icmlaffiliation{betterdata}{Betterdata AI, Singapore}
\icmlaffiliation{SIT}{Singapore Institute of Technology (SIT), Singapore}
\icmlaffiliation{NUS}{National University of Singapore (NUS), Singapore}

\icmlcorrespondingauthor{Milad Abdollahzadeh}{milad@betterdata.ai}

\icmlkeywords{Machine Learning, ICML}

\vskip 0.3in
]



\printAffiliationsAndNotice{}  

\begin{abstract}
Tabular instruction tuning has emerged as a promising research direction for improving LLMs' understanding of tabular data. However, the majority of existing works only consider question-answering and reasoning tasks over tabular data, leaving tabular data generation largely unnoticed.
In this work, for the first time, we explore the efficacy of instruction tuning in improving LLMs' tabular data generation capabilities. 
More specifically, given the high data and computation requirements of tabular instruction tuning, {\bf we aim to address the possibility of instruction tuning for tabular data generation with limited data and computational resources}. 
To achieve this, we first create a high-quality instruction dataset for tabular data, enabling efficient LLM comprehension. We then instruction-tune an open-source LLM (Llama3.1-8B-Instruct) on the training set of this dataset to improve its tabular data generation performance. Our experimental results show that by using our high-quality dataset and instruction-tuning on only 7K instructions with an A100 GPU, for less than 6 hours, {\bf we achieve tabular data generation performance on par with the most capable commercial LLM, GPT-4o}.
\end{abstract}


\section{Introduction}
Large Language Models (LLMs), trained on web-scale corpora, have demonstrated impressive performance across a wide range of natural language processing (NLP) tasks~\cite{Vaswani2017attention, radford2018improving, wang2018glue, hendrycks2021measuring}, and also surprisingly strong performance in following instructions~\cite{wei2022finetuned, ouyang2022training} and reasoning over textual data~\cite{wei2022chain, huang2023towards}. 
These models are widely regarded as emergent repositories of world knowledge~\cite{wei2022emergent, schaeffer2023emergent, roberts2020much}.
However, as their pretraining objectives are inherently optimized for the text modality, which may have some tabular data in the training data, their performance on table-based tasks remains suboptimal~\cite{yang2024unitabe, lin2025ctsyn}.
Recent studies suggest that this limitation stems from the structural mismatch between tabular and textual data: tabular data exhibits a bi-dimensional and relational structure, whereas LLMs are trained using a unidimensional, autoregressive (or masked language modeling) objective, leading to misalignment in inductive biases and representational capacities~\cite{liu2024rethinking, su2024tablegpt2}.

{\bf Tabular Instruction Tuning} has recently emerged as a promising research direction, drawing inspiration from the success of instruction tuning in enhancing the capability of LLMs to handle novel tasks~\cite{ouyang2022training, zhang2023instruction}.
Specifically, recent works~\cite{zhang2024tablellm, zhang2024tablellama, deng2025tama} have proposed generating natural language instructions based on tabular data and using these for instruction-tuning LLMs on table-related tasks.
Studies show that this approach leads to notable improvements in LLMs’ understanding of tabular structures and their performance on tasks involving structured data, such as table-based reasoning and question answering~\cite{deng2022turl, cheng2021hitab, chen2019tabfact}.

{\bf Research Gap.}
Although several works have explored instruction tuning over tabular data, they primarily focus on question answering (QA) and reasoning tasks~\cite{parikh2020totto, aly2021feverous, zhong2017seq2sql, chen2020hybridqa}. 
\textbf{\em The task of generating tabular data, however, remains largely unaddressed.}
Beyond understanding tabular data, which has been the main focus of prior research, the ability to generate realistic and domain-relevant tabular data is increasingly important, especially given the widespread presence of such data in the scientific community and its critical role across various real-world applications~\cite{van2024position, hollmann2023tabpfn, hollmann2025tabpfnv2}. Enabling LLMs to generate synthetic tabular data can help augment limited real-world datasets and accelerate the adoption of machine learning techniques in data-scarce domains.
This work aims to fill this gap by investigating the effectiveness of instruction tuning for enhancing the tabular data generation capabilities of LLMs.

{\bf Limitations.}
The main limitation in exploring the efficacy of instruction tuning for tabular data generation is the \textbf{\em high requirements for large-scale data and extensive computational resources}.
For example, the recent state-of-the-art model TableLlama uses around \textbf{\em 2 million tabular instructions} and \textbf{\em 48 A100 GPUs} to instruction-tune the base LLM and improve its performance on table-based question answering and reasoning tasks.

{\bf In this paper,} we aim to answer the following question: \textbf{\em Can we improve the tabular data generation capabilities of LLMs by instruction tuning these models on limited data and with a limited amount of compute?}
\\
To answer this question, we first create a high-quality instruction dataset for conditional tabular data generation, including 10K instructions. This dataset is gathered from various domains, and extensive metadata is included, together with a snapshot of the input table, to help the LLM follow the context better. We then fine-tune an open-source LLM on this instruction dataset using a single A100 GPU (for less than 6 hours). We show that this instruction-tuning on a limited but high-quality dataset can significantly increase the base LLM's capability in tabular data generation with competitive results compared to the most capable commercial LLM, GPT-4o.
Our main contributions are:
\begin{itemize}
    \item To the best of our knowledge, for the first time in the literature, we explore the efficacy of instruction tuning on improving the performance of the LLMs for tabular data generation.
    \item We create a high-quality instruction dataset for the tabular data generation task to steer the LLM to more precise tabular data generation by including the general and column-wise description of the table as metadata.
    \item Experimental results show that instruction tuning with limited resources and on this limited but high-quality instruction dataset can considerably improve the performance of the base LLM on tabular data generation, and deliver a performance on par with powerful models like GPT-4o.
\end{itemize}


\section{Related Work}
{\bf Tabular Instruction Tuning.}
TableLLM \cite{zhang2024tablellm} performs tabular instruction tuning on LLMs to enable handling various operations on tabular data with LLMs like QA, and Pandas code generation for visualization and analysis purposes.
TableLlama \cite{zhang2024tablellama} creates a large instruction dataset for table-based QA and reasoning tasks and instruction-tunes LLM on this dataset.
TAMA \cite{deng2025tama} analyzes the impact of hyperparameter selection on efficient tabular instruction tuning.
However, none of these works addresses the tabular data generation task with instruction tuning.

{\bf Tabular Data Generation.}
Before the emergence of LLMs, generative models like GANs~\cite{zhao2021ctab, zhao2024ctab, abdollahzadeh2023survey}, VAEs~\cite{wang2025ttvae}, and Diffusion Models~\cite{shi2025tabdiff} were the primary methods for generating tabular data. 
Recently, leveraging LLMs' strong text generation capabilities, multiple works have focused on converting tabular data into text and then fine-tuning LLMs for tabular data generation~\cite{borisov2022language, zhao2023tabula, wang2024harmonic}. However, these models often struggle to follow table-based instructions~\cite{zhang2024tablellama, zhang2024tablellm}.


\section{Problem Setup}
Let $\mathcal{T}$ denote a table with $\mathcal{R}$ rows and $\mathcal{C}$ columns, and $\mathcal{M}$ represent its associated metadata (e.g., table title, description). 
The objective of the instruction following for the tabular data generation with an LLM $f_{\theta}$ is to generate a new table $\mathcal{T}'$. This generation is conditioned on the input table $\mathcal{T}$, its metadata $\mathcal{M}$, and an instruction $\mathcal{I}$ describing the desired generation task for $\mathcal{T}'$:
\vspace*{-1mm}
\begin{equation}
    f_{\theta} (\mathcal{I}, \mathcal{T}, \mathcal{M}) \rightarrow \mathcal{T}'
\end{equation}
Ideally, $\mathcal{T}'$ should follow the distribution of $\mathcal{T}$. This means $\mathcal{T}'$ should possess the same column structure (features) as $\mathcal{T}$ and preserve both the intra-column distributions for each column and the inter-column relationships observed in $\mathcal{T}$.

\begin{figure}[t!]
\begin{center}
\centerline{\includegraphics[ width=\linewidth]{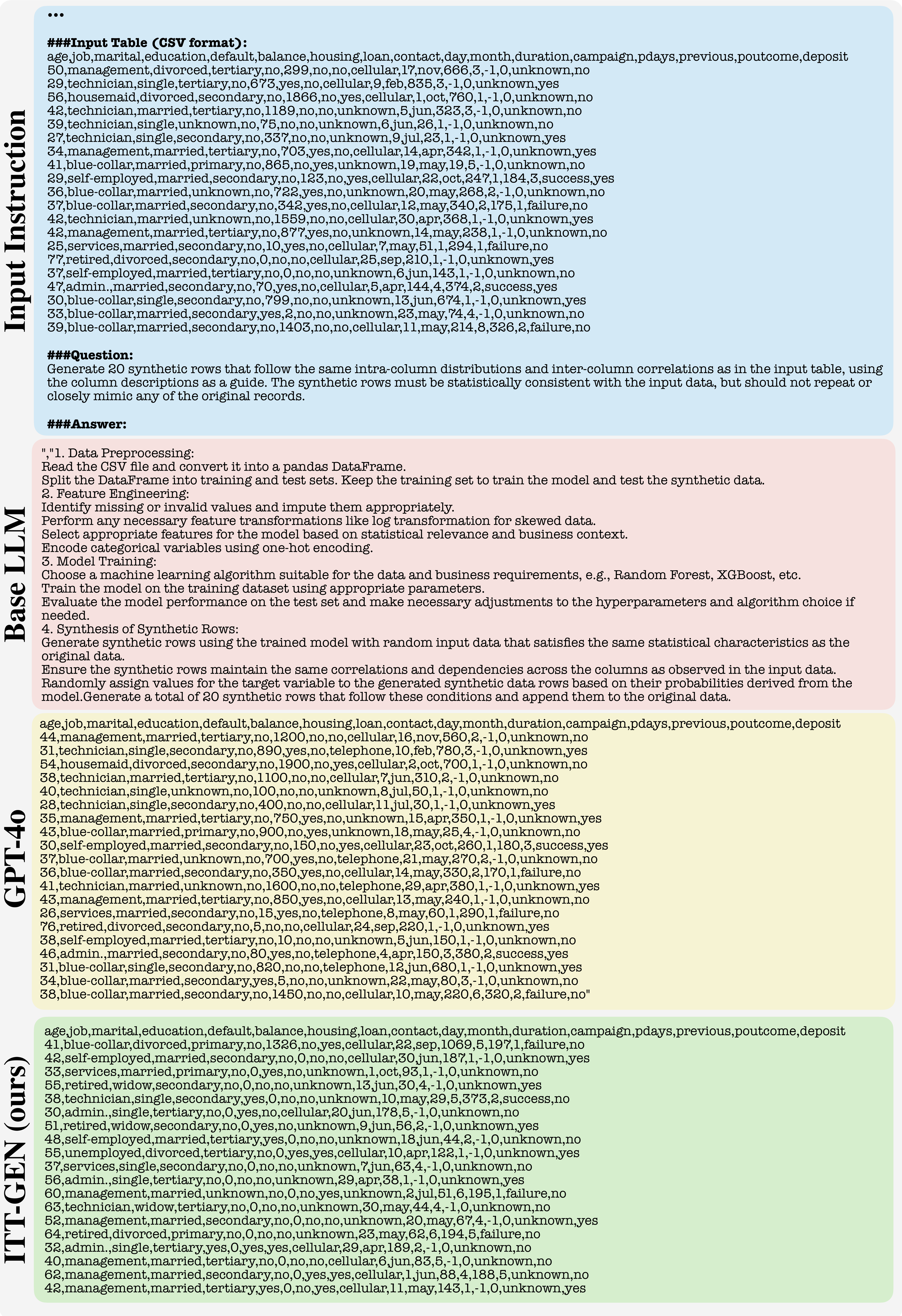}}
\caption{Example of output response of different LLMs for our instruction for tabular data generation.
Base LLM generated some unrelated instructions. However, GPT-4o and our proposed \textsc{ITT-Gen} produce 20 rows of tabular data that follow the same structure, and also the distribution of the input table. Only a part of the input instruction is included due to space limitations. Better viewed when zoomed in.}
\label{fig:llm_response}
\end{center}
\vspace{-2ex}
\end{figure}

\section{Proposed Method}
\label{sec:proposed_mehtod}
In this section, we propose our {\bf I}nstruction {\bf T}uning for {\bf T}abular data {\bf Gen}eration ({\bf \textsc{ITT-Gen}}). 
To improve tabular data generation with LLMs, we perform two main steps: first, we create an instruction dataset for tabular data generation; and next, we fine-tune an open-source LLM on these instructions. In what follows, we discuss the details.

\subsection{Creating Instruction Dataset for Tabular Data Generation}
\label{ssec:instruction_construction}

{\bf Data Collection.}
We sample 20 publicly available tabular datasets that cover 10 different topics. We separate them and select 14 tables for training and in-domain evaluation, and the remaining 6 tables as held-out unseen datasets for out-of-domain (OoD) evaluation. 
The list of these datasets with their topics is shown in Supp., Section~\ref{ssec:supp_dataset} (Table~\ref{tab:dataset}).

{\bf Creating Instruction Dataset.}
For each dataset, in our training set, 
we construct 500 training instances and 100 evaluation instances.
For evaluation datasets, we only construct 100 evaluation instances. 
Each instance in our instruction dataset includes an instruction $\mathcal{I}$ which describes the generation task, an input table $\mathcal{T}$ and its metadata $\mathcal{M}$, and the expected output table $\mathcal{T}'$. The details of constructing each part are as follows:
\begin{itemize}
    \item We manually design the instruction $\mathcal{I}$ to describe the tabular data generation task.
    \item The metadata $\mathcal{M}$ of each table consists of a general description of the table (topic, the general structure, and the applications), and a column-wise detailed description that includes column name, the data types (numerical, categorical, or textual) for each column. We obtain metadata of each table by passing the whole table into GPT-4o~\cite{hurst2024gpt4o} and prompting it to generate this information. 
    We manually go through all generated descriptions to ensure their quality and correctness (More details in Supp., Section~\ref{ssec:supp_details_metadata}).

    \item For input and output tables, we randomly select $N$ rows ($N=20$ in our experiments) of the corresponding table. Our empirical results show that using a set of rows as (expected) output during instruction tuning leads to better results compared to the next token prediction used in previous works~\cite{wang2024harmonic}. 
\end{itemize}
\vspace*{-3mm}
An example of the created instruction is shown in Supp., Section~\ref{ssec:supp_example_instruction} (Figure~\ref{fig:instruction}).

\subsection{Instruction-tuning LLM}
\label{ssec:fine-tuning}
After creating the instruction dataset for tabular data generation, we fine-tune an LLM on the training set of this dataset to improve its tabular generation capabilities.
We use Llama3.1-8B-Instruct as our base model. 
This is a compact model from Llama3 herd of models~\cite{grattafiori2024llama}, 
where a post fine-tuning~\cite{rafailov2023direct} is performed on Llama3.1-8B to enhance its textual instruction following behavior.

Note that our approach is agnostic to the choice of base LLM. 
In Supp., Section~\ref{ssec:supp_results_another_basellm}, we provide additional experimental results to show that our approach also improves tabular data generation performance of TableLlama~\cite{zhang2024tablellama} (SOTA open-source model for table understanding tasks) as the base LLM.


\section{Experiments}

\subsection{Experimental Setup}

{\bf Details of Training and Inference.}
As mentioned in Section~\ref{ssec:fine-tuning}, in our experiments, we used Llama3.1-8B-Instruct \cite{grattafiori2024llama} as our base model. 
We fine-tuned Llama3.1-8B-Instruct on our proposed instruction dataset for tabular data generation with the Huggingface transformers library~\cite{wolf2020transformers}. 
Considering that we have 500 instructions for each of the 14 datasets used for training, we mixed all these 7000 instructions and randomly shuffle them.
We used a learning rate of 2e-5 with a batch size of 3. We trained our model on an A100 80GB GPU for 2 epochs.
We employed DeepSeed training with ZeRO-2 stage~\cite{Rajbhandari2020Zero} for more efficient training.

{\bf Models for Comparison.}
To the best of our knowledge, there are no similar works in the literature that perform instruction tuning for tabular data generation. Therefore, we compare our proposed model with two models: i) Llama3.1-8B-Instruct~\cite{grattafiori2024llama} as the base LLM used in this study, and ii) GPT-4o~\cite{hurst2024gpt4o}, which is one of the most capable commercial LLMs at the time of writing this paper~\cite{shahriar2024putting}.

{\bf Evaluation Metrics.}
We follow the existing tabular data generation works~\cite{zhao2021ctab, zhao2024ctab, li2025tabtreeformer, shi2025tabdiff} and evaluate our approach using fidelity and utility metrics. The details are as follows:
\begin{itemize}
    \item {\bf Fidelity} measures the distributional similarity between generated and tabular data. Two well-known metrics for measuring fidelity of the generated data are: {\em i) Shape,} which measures the similarity between the marginal distribution of the real and generated data for each column~\cite{zhang2024mixedtype}, and {\em ii) Trend} which measures the capability of the generated data to capture the correlation between different columns \cite{shi2025tabdiff}.
    Higher values of {\em Shape} and {\em Trend} metrics indicate a higher data fidelity.

    \item {\bf  Utility} evaluates whether generated tabular data is useful for a downstream task.
    To evaluate the utility, we use Train-on-Synthetic, Test-on-Real ({\em TSTR}) framework~\cite{xu2019modeling}, which trains the model on generated (synthetic) tabular data, and then performs the evaluation on held-out real tabular data.
    For this framework, we use three different models (for training and evaluation), including linear, random forest~\cite{breiman2001random}, and XGBoost (XGB)~\cite{chen2016xgboost}.
\end{itemize}

\subsection{Experimental Results}
\label{ssec:rexperimental_results}
An example of generated output for our input instruction is shown in Figure~\ref{fig:llm_response}. As one can see, our proposed \textsc{ITT-Gen} and GPT-4o are able to generate tabular data that follows the same distribution as the input table. However, base LLM (Llama3.1-8B-Instruct) fails to follow our instructions to generate tabular data and starts to generate some irrelevant instructions. We remark that a similar behavior happens for most of the instructions, and only for some instructions, the base LLM is able to generate limited rows (not the whole 20 rows asked) of tabular data. Nevertheless, we collect all generated tabular data and filter out the irrelevant parts to be able to report fidelity and utility metrics for the base LLM.

\begin{table}[h]
\centering
\caption{Fidelity results across different algorithms.}
\vspace{1.5ex}
\resizebox{\columnwidth}{!}{%
\begin{tabular}{lcccccc}
\toprule
\textbf{Dataset} & \multicolumn{2}{c}{\textbf{Base LLM}} & \multicolumn{2}{c}{\textbf{\textsc{ITT-Gen (ours)}}} & \multicolumn{2}{c}{\textbf{GPT-4o}} \\
\cmidrule(lr){2-3} \cmidrule(lr){4-5} \cmidrule(lr){6-7}
  & Shape & Trends & Shape & Trends & Shape & Trends \\
  
\midrule
adult  & 87.48 & 75.13 & 85.73 & 52.54 & 92.34 & 87.96 \\
bank  & 75.63 & 65.08 & 85.57 & 86.34 & 93.42 & 91.7 \\
bestseller  & 89.12 & 90.5 & 89.56 & 93.16 & - & 86.4 \\
biodeg  & 89.59 & 80.04 & 91.68 & 86.61 & 94.12 & 86.54 \\
boston  & 88.91 & 87.47 & 92.38 & 88.98 & 90.87 & 93.02 \\
breast\_cancer  & 55.31 & 37.07 & 84.12 & 69.36 & 78.65 & 64.16 \\
BTC-USD\_stock  & 90.19 & 95.06 & 88.2 & 99.31 & 93.52 & 98 \\
california\_housing & 88.7 & 90.52 & 73.29 & 80.06 & 96.27 & 97.84 \\
car\_prediction\_data & 74.17 & 54.44 & 84.59 & 60.77 & 78.8 & 61.97 \\
credit-g & 88.3 & 78.38 & 86.29 & 75.05 & 93.12 & 86.67 \\
diabetes  & 89.45 & 91.02 & 83.41 & 88.77 & 89.93 & 88.11 \\
healthcare\_insurance & 88.52 & 74.35 & 91.76 & 86.74 & 93.14 & 88.39 \\
iris & 82.69 & 55.39 & 88.17 & 77.86 & 89.58 & 87.13 \\
job\_posting  & 40.52 & 22.4 & 54.55 & 36.15 & 64.56 & 41.01\\
Players2024  & 34.84 & 11.48 & 53.55 & 16.69 & 53.09 & 16.13 \\
room\_occupancy & 81.56 & 74.99 & 86.89 & 81.56 & 88.42 & 91.11 \\
supermarket\_store\_branches & 90.36 & 97.88 & 83.2 & 90.45 & 93.85 & 96.46 \\
tour\_travels\_customer\_churn & 84.33 & 70.19 & 91.59 & 75.18 & 90.69 & 75.86 \\
twitter\_astrazeneca\_anti\_covid & 84.7 & 96.47 & 91.83 & 98.65 & 75.67 & 98.03 \\
wdbc & 85.89 & 88.26 & 87.43 & 92.62 & 90.21 & 96.08 \\
\bottomrule
\end{tabular}%
}
\label{table:all_fidelity}
\end{table}

{\bf Fidelity Results.}
Table~\ref{table:all_fidelity} shows the fidelity results for generated tabular data with different algorithms. As one can see, the proposed \textsc{ITT-Gen} approach has on-par performance with the powerful GPT-4o model. 
Note that for base LLM, even though the metrics show competitive performance, these are calculated only for the portion of the output that is tabular data (~20\%), and the remaining unrelvenet generated data (~80\% of generated output with base LLM) is discarded for the sake of only being able to report these metrics.

{\bf Utility Results.}
Table~\ref{table:ml_utility} shows the utility results for different approaches. Similarly, the proposed \textsc{ITT-Gen} yields a performance on par with GPT-4o indicating that generated tabular data with our instruction-tuned LLM can be efficiently used for downstream tabular tasks.

\begin{table}[t]
\centering
\caption{Utility result for synthetic data.  Averaged AUC and R2 scores are reported for classification and regression datasets, respectively. `--' indicate the output can not be used to train an ML model. Note that we only report a subset of datasets here. Others follow the same trend.}
\vspace{1.5ex}
\resizebox{\columnwidth}{!}{%
\begin{tabular}{l c c c c}
\hline
\textbf{Dataset} &  \textbf{Real} & \textbf{BaseLLM} & \textbf{\textsc{ITT-Gen}} & \textbf{GPT-4o}\\
\hline
adult & 0.8796 & 0.655867 & 0.826533 & 0.873200 \\
bank & 0.800720 & 0.353441 & 0.616246 & 0.819928 \\
bestseller & 0.781972 & --  & 0.743701 & 0.710766 \\
biodeg & 0.917188 & 0.816096 & 0.862471 &  0.922341 \\
boston & 0.745258 & 0.677436 & 0.655484 &  0.729943 \\
berast\_cancer & 0.9942 & -- & 0.9831 & 0.9919 \\
BTC-USD-stock & 0.995497 & 0.917406 & 0.993921 & 0.990918  \\
California housing & 0.640855 & 0.393930  & 0.497865 & 0.589859 \\
Diabetes & 0.82038 & 0.821207 & 0.798160 & 0.797334  \\
Healthcare insurance & 0.737844 & 0.360006 & 0.695602 & 0.716192 \\
Iris & 1.0000 & -- & 0.987143 & 0.997149 \\
Players 2024 & 0.380000 & 0.327586 & 0.425532 & 0.464481 \\
Room Occupancy & 0.993658 & 0.976697 & 0.993144  & 0.994749 \\
Tour \& Travels Cusomer Chorn &  0.767578 &  0.685234 & 0.543672  & 0.706484 \\
Twitter Atrazenca Anti Covid & 0.9457 & 0.89584 & 0.93094 &  0.93573 \\
Wdbc & 0.99235 & 0.982966 & 0.979396 & 0.988066 \\

\hline
\end{tabular}
} 
\label{table:ml_utility}
\end{table}

\section{Conclusion}
In this paper, for the first time in the literature, we explore the potential of leveraging instruction tuning to improve tabular data generation performance. For this, we create an instruction dataset for the tabular data generation task, and instruction-tune an open-source base LLM on this dataset.
Our results suggest that instruction-tuning on our small but high-quality dataset with only one A100 GPU and for less then 6 hours, can yield a performance on par with GPT-4o, the most capable commercial LLM.

\nocite{langley00}

\bibliography{references}
\bibliographystyle{icml2025}

\newpage
\appendix
\onecolumn

\section{Additional Details on Our Instruction Dataset}

\subsection{Details of the Datasets}
\label{ssec:supp_dataset}
Table~\ref{tab:dataset} tabulates the details of the public datasets used to create our instruction dataset for tabular data generation.

\begin{table*}[ht]
\caption{We sample 20 publicly available datasets to create our instruction dataset for tabular data generation. To ensure diversity, these datasets are sampled from 10 different topics. For each dataset (table), $\mathcal{R}$ and $\mathcal{C}$ denote the number of rows (samples) and the number of columns (features), respectively. \textsc{Train} indicates whether a dataset is used during training.}
\label{tab:dataset}
\vspace{1.5ex}
\begin{center}
\begin{small}
\begin{sc}
\resizebox{\linewidth}{!}{
\begin{tabular}{lcccr}
\toprule
Topic & Dataset & $\mathcal{R}$ & $\mathcal{C}$ & Train  \\
\midrule
\multirow{4}{*}{\textbf{Consumer and Market Analysis}}  & Amazon Top 50 Bestselling Books (2009-2019) & 550 & 7 & \cmark \\
    & Bitcoin BTC-USD Stock Dataset
 & 2836 & 7 & \cmark\\
    & Car Price Prediction Dataset & 1000 & 7 & \cmark\\
    & Supermarket Store Branches Sales Analysis & 896 & 5 & \xmark\\
    \arrayrulecolor{black!30}\cmidrule(r){1-5}

    \multirow{4}{*}{\textbf{Healthcare and Medical Research}}  & US Health Insurance Dataset
 & 1338 & 7 & \cmark \\
    & Breast Cancer Wisconsin
 & 699 & 10 & \cmark\\
    & Diabetes
 & 768 & 9 & \cmark\\
    & Wdbc - Breast Cancer Diagnosis & 569 & 31 & \xmark\\
    \arrayrulecolor{black!30}\cmidrule(r){1-5}

\multirow{3}{*}{\textbf{Finance and Credit Risk Analysis}}  & Adult Income (UCI Census Income)
 & 48842 & 15 & \cmark \\
    & Bank Marketing
 & 45211 & 17 & \cmark\\
    & Credit-g
 & 1000 & 21 & \xmark\\
    \arrayrulecolor{black!30}\cmidrule(r){1-5}

\multirow{2}{*}{\textbf{Employment and Workforce Analytics}}  & Football Players Season 2024
 & 5935 & 7 & \cmark \\
    & Job Posting
 & 1095 & 6 & \cmark\\
    \arrayrulecolor{black!30}\cmidrule(r){1-5}

\multirow{2}{*}{\textbf{Real Estate and Housing Economics}}  & Boston Housing
 & 506 & 14 & \cmark \\
    & California Housing
 & 20640 & 10 & \xmark\\
    \arrayrulecolor{black!30}\cmidrule(r){1-5}

\textbf{Energy and Smart Building Systems
}  & Room Occupancy Dataset & 2665 & 6 & \cmark \\
    \arrayrulecolor{black!30}\cmidrule(r){1-5}

\textbf{Transportation and Travel Industry
}  & Tour \& Travels Customer Churn Prediction & 954 & 7 & \cmark \\
    \arrayrulecolor{black!30}\cmidrule(r){1-5}

\textbf{Social Media Analytics}  & Twitter AstraZeneca AntiCovid & 1553 & 5 & \xmark \\
    \arrayrulecolor{black!30}\cmidrule(r){1-5}

\textbf{Chemistry and Environmental Science}  & Qsar-biodeg  & 1055 & 42 & \xmark \\
    \arrayrulecolor{black!30}\cmidrule(r){1-5}

\textbf{General Machine Learning Benchmarks}  & Iris  & 150 & 5 & \cmark \\

\bottomrule
\end{tabular}
}
\end{sc}
\end{small}
\end{center}
\vskip -0.1in
\end{table*}

\subsection{Example of Created Instruction in Our Instruction Dataset}
\label{ssec:supp_example_instruction}
\begin{figure*}[h]
\begin{center}
\centerline{\includegraphics[ width=\linewidth]{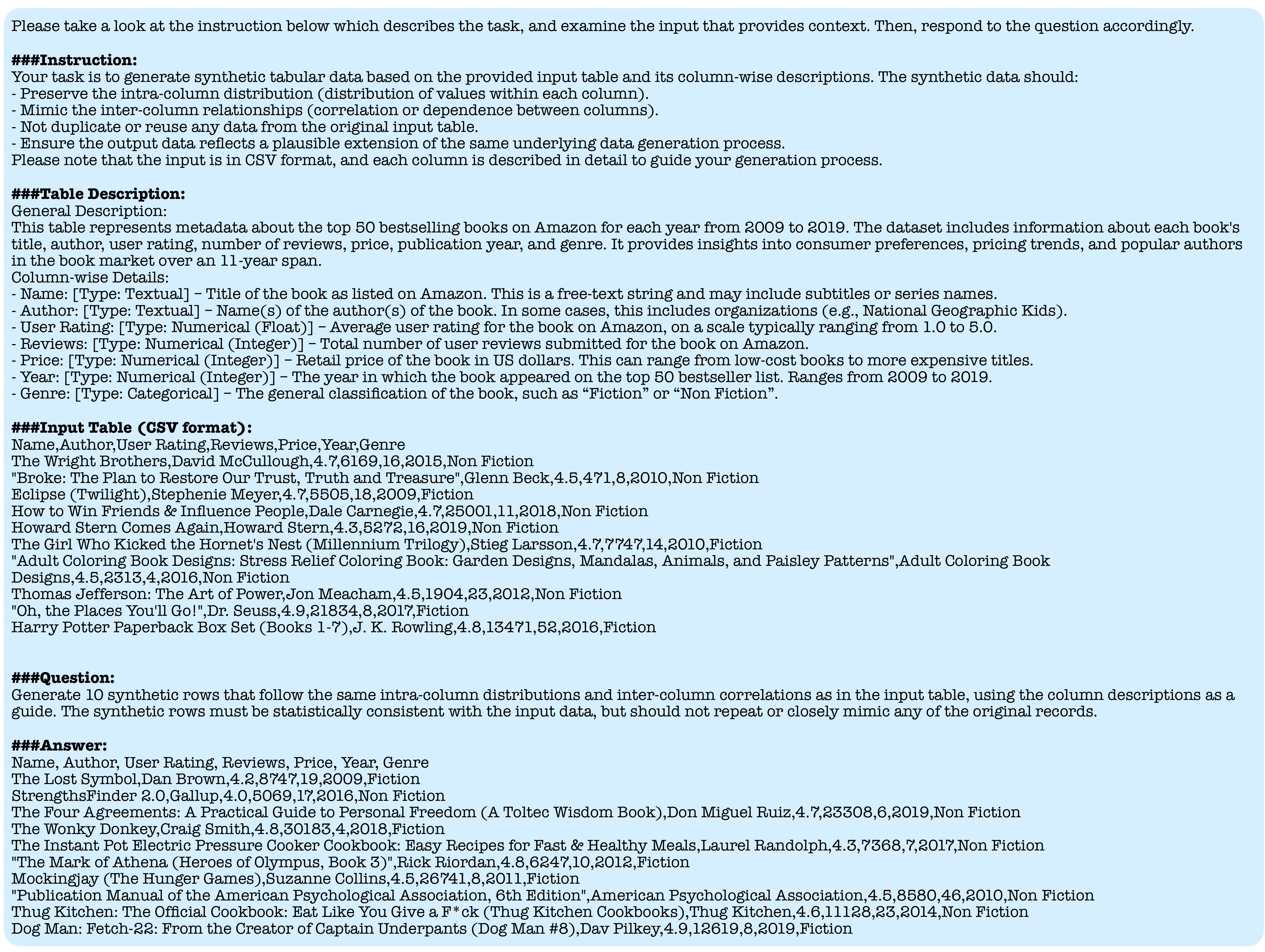}}
\caption{An example of the instruction created in our dataset for tabular data generation.}
\label{fig:instruction}
\end{center}
\vspace{-5ex}
\end{figure*}

An example of a created training instruction is shown in Figure~\ref{fig:instruction}.

\subsection{Details of Metadata Generation for Our Instructions}
\label{ssec:supp_details_metadata}

As mentioned in the main paper, the metadata for each table includes a general description of the table and column-wise details.
Some of the tables lack such metadata, and for some, various descriptions are available online.
Our preliminary experimental results suggest the importance of high-quality metadata in steering LLMs for proper tabular data generation.
Therefore, to ensure the quality of the metadata used in our instructions, we leverage GPT-4o for metadata generation.

\begin{figure}[h!]
\begin{center}
\centerline{\includegraphics[width=\linewidth]{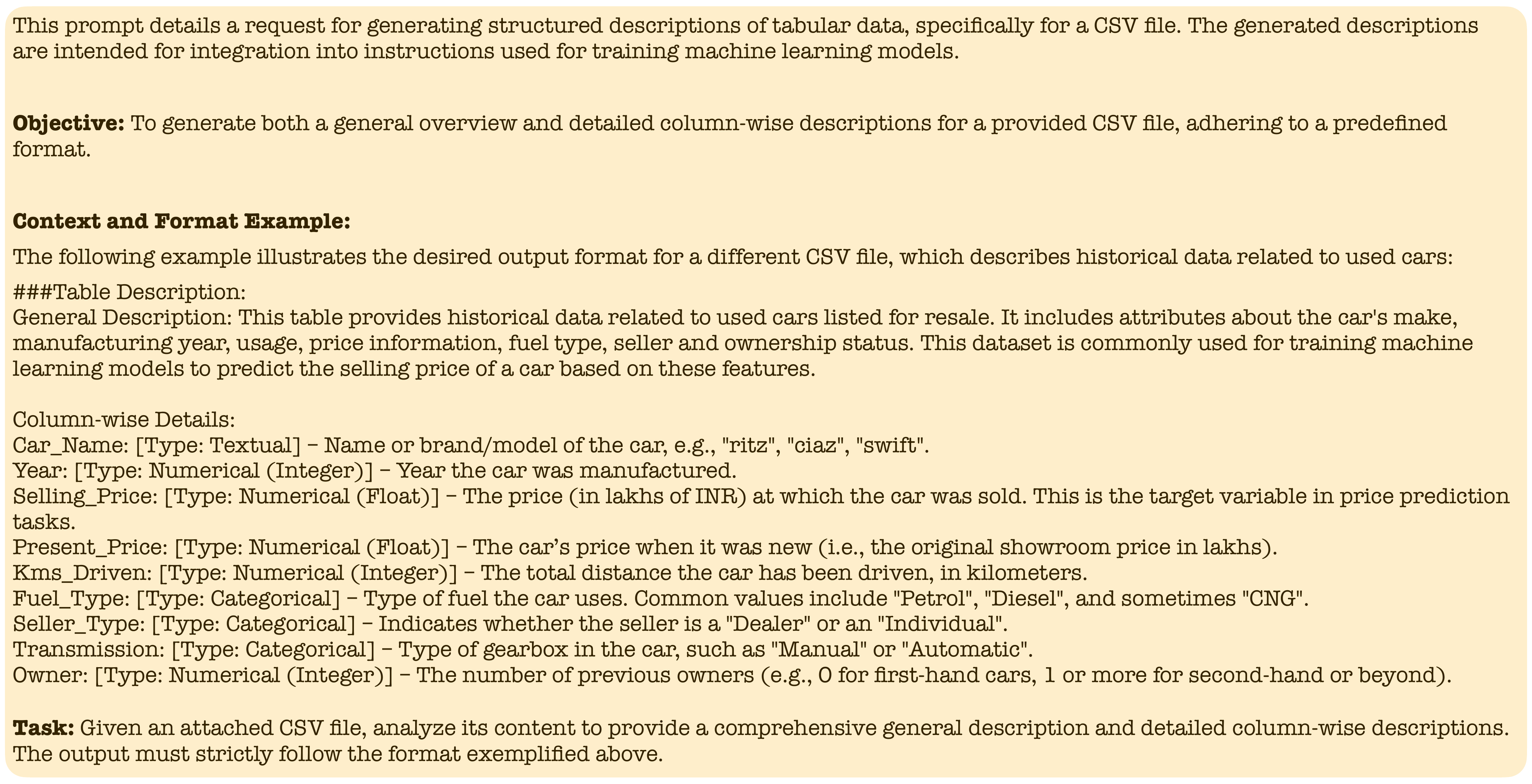}}
\caption{Template used to prompt GPT-4o for generating table descriptions.}
\label{fig:template_description_generation}
\end{center}
\vspace{-5ex}
\end{figure}

Specifically, to unify the format of the descriptions and ensure that all required details (e.g., column name, column data type, etc.) are present in the generated description, we manually extract the general and column-wise descriptions for one of the tables.
We then use this as context and design a template prompt as input to GPT-4o. This template prompt is shown in Figure~\ref{fig:template_description_generation}, and it is used to obtain the table descriptions for all tables.
After obtaining these descriptions from GPT-4o, we review all generated descriptions to ensure their quality and accuracy.

\newpage

\section{Additional Experimental Results}
In this section, we include additional experimental results that could not be presented in the main submission due to space limitations.
To demonstrate that our proposed approach is model-agnostic, we also include instruction-tuning results with another base LLM, TableLlama~\cite{zhang2024tablellama}.

\subsection{Instruction-tuning with Another Base LLM}
\label{ssec:supp_results_another_basellm}
In this section, we provide additional experimental results using TableLlama~\cite{zhang2024tablellama} as our base LLM.
TableLlama is pre-trained on a variety of table-based tasks, including question answering, reasoning, table fact verification, and table-to-text generation.
It is considered a state-of-the-art open-source LLM for table-based tasks, outperforming GPT-3.5 and demonstrating competitive performance compared to GPT-4.
TableLlama is obtained by fine-tuning LongLoRA 7B~\cite{chen2024longlora} on 3M table-based Q\&A and reasoning instructions.
Note that LongLoRA 7B itself is derived from Llama 2~\cite{touvron2023llama2openfoundation} by replacing vanilla attention with shift short attention, thereby increasing the context window size to 8192 tokens.
We fine-tune TableLlama on our proposed instruction dataset for conditional generation using the Huggingface Transformers library~\cite{wolf2020transformers}.

The results of instruction-tuning TableLlama on our dataset for tabular data generation are shown in Table~\ref{table:fidelity_tablellama} for the fidelity metric and in Table~\ref{table:ml_utility_tablellama} for the utility metric.
As one can see, the base LLM does not perform well on tabular data generation, even though it is trained on a large set of table-based tasks.
In fact, these results emphasize the point that tabular data generation is a distinct task compared to Q\&A and reasoning.
However, after fine-tuning, the model's performance improves significantly in terms of both the fidelity and utility of the generated responses. This illustrates that the base LLM struggles to generate meaningful output in the context of tabular data generation, but after instruction tuning, the generated data improves significantly. It better follows the structure of the tabular data, and the output more closely mimics the intra-column distributions and inter-column relationships. Note that since the base LLM used in TableLlama (Llama2) is relatively outdated, even after instruction tuning, there remains a considerable performance gap compared to a strong commercial model like GPT-4o, which is trained on far more tokens and has significantly higher capacity.

\begin{table}[h]
\vspace{-2ex}
\centering
\caption{Fidelity result for synthetic data using TableLlama~\cite{zhang2024tablellama} as base LLM for our instruction tuning. 
Note that `--' indicates that the output of the base LLM (TableLlama) does not follow the structure of the tabular data, and therefore can not be used for fidelity calculation.
}
\vspace{1.5ex}
\resizebox{0.35\columnwidth}{!}{%
\begin{tabular}{llcc}
\toprule
\textbf{Dataset} & \textbf{Algorithm} & \textbf{Shape} & \textbf{Trends} \\
\midrule
\multirow{3}{*}{California} & TableLlama & -- & -- \\ 
& ITT-\textsc{Gen} (Ours) & 78.57 & 79.75 \\
 & GPT-4o & 94.8 & 86.55 \\
\cline{1-4}

\multirow{3}{*}{Credit} & TableLlama & -- & --\\
& ITT-\textsc{Gen} (Ours) & 60.23 & 37.25 \\
 & GPT-4o & 90.99 & 80.15 \\
\cline{1-4}

\multirow{3}{*}{Boston} & TableLlama & -- & --
\\ & ITT-\textsc{Gen} (Ours) & 75.84 & 75.63 \\
 & GPT-4o & 89.92 & 89.63 \\
\cline{1-4}

\multirow{3}{*}{Diabetes} & TableLlama & -- & --
\\ & ITT-\textsc{Gen} & 66.14 & 70.9 \\
 & GPT-4o & 92.1 & 91.15 \\
\bottomrule
\end{tabular}
}
\label{table:fidelity_tablellama}
\vspace{-3ex}
\end{table}

\begin{table}[h]
\centering
\caption{Utility result for synthetic data using TableLlama~\cite{zhang2024tablellama} as base LLM for our instruction tuning.
Average AUC and MAPE are reported as utility metrics. Note that `--' indicates that the output of the base LLM (TableLlama) can not be used to train a machine learning model on tabular data.}
\vspace{1.5ex}
\resizebox{0.35\columnwidth}{!}{%
\begin{tabular}{c  c c c}
\hline
\textbf{Dataset} & \textbf{TableLlama} & \textbf{GPT4o} & \textbf{Ours}\\
\hline
Boston ($\downarrow$) &--&0.187&0.257 \\
\hline
California ($\downarrow$) &--&0.334&0.428 \\
\hline
Credit ($\uparrow$) &--&0.767&0.487 \\
\hline
Diabetes ($\uparrow$) &--&0.773&0.721 \\
\hline
\end{tabular}
} 
\label{table:ml_utility_tablellama}
\vspace{-14ex}
\end{table}

\end{document}